# Cross-lingual Keyword Assignment

Ralf Steinberger

European Commission
Joint Research Centre
Institute for the Protection and Security of the Citizen (IPSC)
T.P. 361, 21020 Ispra (VA), Italy
Ralf.Steinberger@jrc.it

**Abstract:** This paper presents a language-independent approach to controlled vocabulary keyword assignment using the EUROVOC thesaurus. Due to the multilingual nature of EUROVOC, the keywords for a document written in one language can be displayed in all eleven official European Union languages. The mapping of documents written in different languages to the same multilingual thesaurus furthermore allows cross-language document comparison. The assignment of the controlled vocabulary thesaurus descriptors is achieved by applying a statistical method that uses a collection of manually indexed documents to identify, for each thesaurus descriptor, a large number of lemmas that are statistically associated to the descriptor. These associated words are then used during the assignment procedure to identify a ranked list of those EUROVOC terms that are most likely to be good keywords for a given document. The paper also describes the challenges of this task and discusses the achieved results of the fully functional prototype.

## 1. Introduction, Users, Related Work

### 1.1. Introduction

In the last years, many useful NLP tools have been developed and many of them are now even available commercially. Most of these tools are monolingual or *multi-monolingual*, meaning that the software can deal with more than one language, but that the results will always be displayed in the same language as the text. We therefore distinguish these applications from *cross-lingual software*, which is software that helps to transgress the language boundary. Examples for such applications are machine translation and cross-lingual document retrieval, i.e. retrieval using search engines which allow to enter a search term in one language and which also yield results in other languages, usually because the query is translated in one way or another. In our eyes, cross-lingual applications are currently the bottleneck of available NLP tools. To our knowledge, there are no applications that allow comparing documents written in different languages with each other and there are very few which give users a quick overview of the approximate contents of documents written in different languages.

### 1.2. Applications and user groups

In this paper, we present such a cross-lingual application that allows users to see keywords in their own language of documents written in other languages. We achieve this *cross-lingual keyword assignment capacity* by assigning descriptors from the multilingual controlled vocabulary thesaurus EUROVOC, which exists in exact translations in all eleven official European Union (EU) languages (see section 3.1). Once the most suitable thesaurus entries for a text in one language have been identified, the equivalent terms can be displayed in all EU languages.

Besides giving users *cross-lingual information access*, this application also allows automatic cross-lingual document similarity calculation. Our approach to document similarity calculation assumes that documents that share more keywords are more similar to each other than documents that share less or no keywords (Hagman 1999). When texts written in different languages are linked to the same multilingual thesaurus, cross-lingual document comparison becomes possible. As document similarity calculation is the basis for document clustering and for the visualisation of large document collections in document maps (Steinberger et al. 2000, Hagman et al. 2000), the EUROVOC keyword assignment tool allows multilingual document clustering and the visualisation of multilingual document collections.

This application is part of a larger effort of putting together a tool set at the JRC which allows to (a) retrieve potentially relevant documents from a variety of sources, (b) analyse them and extract different information aspects from them, and (c) visualise the contents of individual documents as well as of whole docu-ment collections in a variety of ways. A major focus of our effort is on multilingual and cross-lingual applications.

This work is being carried out because there is a real need in and outside the European Commission to get cross-lingual information access and to assign controlled vocabulary keywords automatically. The interested user group includes several national and international organisations, which regularly have to store and retrieve many documents, especially when they are multilingual text collections.

The organisations that currently use people to index[1] all their documents require very high precision so that it is not expected that our automatic method will replace the current manual procedure. However, they may be interested in combining the automatic method with the manual (human) assignment procedure in order to lower the staff cost and to check the consistency of the results. In organisations who do not employ human indexers, even keywords which are chosen less carefully than the manually identified ones will be welcomed because they are better than no indexing terms at all.

### 1.3. Contents

Section 2 presents a monolingual keyword assignment tool, which is used in the training phase of the process of identifying EUROVOC descriptors. Section 3 describes EUROVOC (3.1) and explains the training (3.2.1) and the assignment phases (3.2.3) of the automatic assignment procedure (3.2). Section 4 discusses the achieved results. Section 5 lists difficulties we encountered, and section 6 provides some implementation details.

### 1.4. Related work

The only similar application we are aware of was developed by Ferber (1997), whose main aim was to use a multilingual thesaurus for the retrieval of English documents using search terms in other languages than English. Ferber trained his associative system on the *titles* of 80.000 bibliographic records, which were manually indexed using the OECD thesaurus. The OECD thesaurus is similar to EUROVOC, with the difference that it is smaller and exists only in four languages. Ferber achieved better recall and precision results than we currently do (see section 4.3 below), but the training data available to us is of a different nature so that we cannot use his approach and we cannot compare our results directly with his. Our training corpus consists of about 3500 rather varied texts whose text length varies between two lines and fifty pages (see section 5).

Ferber's approach is different from ours in that he uses the absolute number of occurrences of associated words instead of their keyness, and he uses a variation of the expected mutual information measure to downplay the impact of high-frequency words. Ferber does not consider the hierarchical structure of the thesaurus used.

## 2. *Monolingual keyword assignment using an open set of keywords*

Our monolingual keyword assignment tool, which is needed during the training phase of the EUROVOC descriptor assignment process, chooses a small number of particularly characteristic words from a text. This is different from the EUROVOC descriptor assignment tool described in section 3, which identifies terms from a closed list of terms that often do not occur explicitly in the text.

### 2.1. Functionality of the tool

The tool used at the JRC is a purely statistical tool which compares a word frequency table for a given text with a word frequency table of a reference corpus (see 3.2.2) and identifies the most *characteristic* words of the text, using a choice of standard statistical techniques. In other words: if a certain word occurs significantly more often in a given text than it occurs, on average, in a large selection of 'normal' texts (the reference corpus), this word is identified as a keyword. The tool is a customised standalone version of the keyword identification functionality in the linguistic tool set *WordSmith Tools* (Scott 1999).

The statistical tool currently uses two alternative tests to identify the most characteristic words of a text: the chi-square test and the log-likelihood test. We mainly use the latter algorithm because it works better for low-frequency words (Kilgariff 1996). The tool produces a list of the most significant words of a text, plus an indication of their importance as content descriptors for this document. We refer to this in-dicator as the *keyness* of the keyword. **Table 1** shows the list of keywords assigned automatically for the document you are currently reading.

---

[1] In this paper, we consistently use the verb 'to index' in the sense 'to assign keywords' (i.e. a small number of words or multi-word terms which represent the contents of a document) and never in the sense 'full-text indexing' (i.e. to produce an inverted index of all words occurring in a document).

The tool produces rather accurate lists of keywords. Not all suggested keywords are highly meaningful, but they usually provide users with a rather good idea of the document contents. The tool can be applied to any language. All the input it needs is a word frequency list of a reference corpus and a lemmatiser or stemmer for the languages.

## 2.2. Minimal linguistic input

As function words and other high-frequency words are not meaningful as keywords even if they are much more frequent than in 'normal' text, they can be excluded by being inserted into an extensive list of ***stop words***. This stop word list is also used to exclude content words, which are not particularly relevant within a certain subject domain. For instance, in a homogeneous collection of purely internet-related documents, the words *internet*, *web*, *http*, etc. are clearly not wanted as keywords as they are the terms all documents have in common.

For highly inflectional languages such as Spanish, German or French, statistics based on word *forms* are not very significant because they ignore the relationship between the words of the same morphological paradigm. Even for the English language, it is advised to make explicit the link between singular and plural forms of nouns and between the few verb form alternations. Therefore, we apply **lemmatising software**[2] which converts all word forms in running text to their base form, also referred to as *dictionary form*. Both the texts for which keywords should be identified and the whole of the reference corpus are lemmatised before the word frequency tables are compared. This has the desired consequence that the frequency of the *lemma* is considered instead of the frequency of the more arbitrary word form, so that the keyword identification tool only suggests lemmas as keywords.

The tool only considers single words so that words, which are part of **multi-word expressions**, are not considered in their context. For a text containing several occurrences of the multi-word term *power plant*, for instance, the tool may identify one or both of the single words *power* and *plant* as keywords, but without context, at least the word *plant* is misleading. When applying the tool to a text collection pertaining to a specific sublanguage, we therefore identify lists of the most frequent multi-word expressions[3] (currently the approximately 500 most frequent ones) and mark these up in both the reference corpus and the documents, using the underscore (*power_plant*). This solution is clearly not very elegant and the multi-word expressions considered are limited to this pre-established list, but it improves the performance.

**Table 1** Automatically identified keywords for this document, plus their keyness.

| Keywords | Keyness |
|---|---|
| descriptor | 1478.40 |
| eurovoc | 699.25 |
| keyword | 557.60 |
| text | 412.42 |
| document | 409.63 |
| assignment | 312.67 |
| list | 283.57 |
| thesaurus | 253.45 |
| associate | 237.49 |
| tool | 235.56 |
| cross-lingual | 233.03 |
| manually | 191.30 |
| language | 190.35 |
| word | 189.93 |
| corpus | 164.26 |
| multilingual | 150.46 |
| lemma | 137.40 |
| multi-word | 134.62 |
| training | 123.15 |
| term | 120.22 |
| index | 116.99 |
| keyness | 116.51 |
| assign | 111.12 |
| … | |

## 3. Cross-lingual keyword assignment using a controlled vocabulary

The monolingual keyword assignment tool presented in section 2 does not make any conceptual abstraction whatsoever because it can only suggest keywords which actually occur in the text. For instance, the keyword lists for one text talking about 'bread' and for another text talking about 'toast' will contain the keywords

---

[2] The lemmatising software used is the *IntelliScope Search Enhancer*, version 2.0, by *Lernout & Hauspie*. This software, which can also normalise regional spelling variations, date and currency expressions, etc., was chosen because it has the advantage over alternative products that it is available for a wide variety of languages.

[3] The *Text Analysis Tool* of the *Euramis Client Interface*, which we use for the recognition of multi-word terms, was developed by the European Commission's in-house *Translation Service* in Luxembourg.

*bread* and *toast*, respectively, so that it is up to the user to make the conceptual link between these two types of *bakery products*. When the keyword lists are used as input for automatic document comparison and clustering (see section 1.2), the lack of conceptual abstraction is clearly a disadvantage because the clustering program only compares strings (keyword lists) and will therefore not be able to consider the link between both semantically closely related nouns.

One way of overcoming this lack of abstraction is to choose the keywords from a restricted (controlled) list of thesaurus terms. We chose to use the EUROVOC thesaurus (EUROVOC, 1995) because it exists in exact one-to-one translations in all official European Union languages and because we got access to manually keyword-assigned text collections which we could use as training material.

### 3.1. EUROVOC thesaurus

EUROVOC is a multilingual thesaurus, which exists, in exact one-to-one translations in all eleven official EU languages. It consists of 5.933 *descriptors* (keywords) which are ordered in a hierarchical structure with a maximum of eight levels. At the top level, there are 21 categories, called *fields*, and at the level below there are 127 categories, referred to as *microthesauri*. There are 5877 reciprocal relations linking *broader terms* (BTs; superordinates) and *narrower terms* (NTs; subordinates) with each other. 2.730 reciprocal associations mark *related terms* (RT) from different parts of the thesaurus structure. There is also a language-dependent number of descriptor synonyms which are related to the descriptor by the relation *use for* (UF).

Eurovoc was developed for use by the archivists of the *European Parliament* (EP), the *European Commission*'s *Publications Office* (OPOCE), and many national organisations as a controlled vocabulary to index all documents in the archives manually.

### 3.2. Assignment procedure

EUROVOC thesaurus terms (referred to as *descriptors*) typically are rather abstract multi-word expressions such as ORGANISATION OF ELECTIONS, POLITICAL IDEOLOGY and FORESTRY ECONOMICS.[4] Texts covering these subjects are unlikely to contain these exact terms so that searching for the EUROVOC descriptor strings in text is not an option (see also section 4.2). Instead, we produce large lists of related words, which are semantically or otherwise associated to the EUROVOC descriptors. We refer to these single words as *associates*. The basic idea is that, when large amounts of associates for a certain descriptor are found in a new text, the chances are good that this descriptor is a suitable keyword for the text. The manual identification of the associates for 5.933 descriptors in many different languages would be a very tedious and laborious task. We therefore developed a system which identifies these associates automatically, using manually indexed training material (training phase, 3.2.1), and which assigns EUROVOC descriptors to a new text with a certain probability when many associates for this descriptor were found in this text (assignment phase, 3.2.3).

#### 3.2.1. Training Phase

The goal of the training phase is to identify a ranked list of associates for each descriptor in each language, where each associate has a certain weight, which indicates its degree of association with the descriptor. As the lists of associates are typically rather long, it is crucial that the weight of each associate is calculated so that it can be used in the assignment procedure.

We produce the lists of descriptor associates by selecting all those documents of one language that were manually indexed with the first descriptor, and by producing a mega-document with this selection. This mega-document is then treated as one document and is subjected to the monolingual keyword identification tool presented in section 2. The outcome of this procedure is a list of the most characteristic words of this mega-document, and hence of the most characteristic associates for this first descriptor. The same method is repeated for all descriptors for which enough training material exists. We consider two to three pages of training material the bare minimum to guarantee associate lists of reasonable quality. Currently, we have enough training material for 2870 descriptors. The average number of manually assigned descriptors per text in our Spanish training corpus is 6.91.

**Table 2** shows the first part of the Spanish list of associates for the EUROVOC descriptor GESTIÓN DE LA PESCA (FISHERY MANAGEMENT). The list shows that the associates are mostly related semantically to the descriptor term (most words pertain to the semantic fields of *fishery*/*pesca* and *management/gestión*), but it also includes geographic expressions and related acronyms. Although not all of the associated words are part of the same semantic field, they are all in a statistical co-occurrence relation with the descriptor.

---
[4] We shall adhere to the convention of writing EUROVOC descriptors in SMALL-CAPS.

**Table 2** Spanish Associates of the Eurovoc descriptor GESTIÓN DE LA PESCA (FISHERY MANAGEMENT)

| Associate | Weight |
|---|---|
| pesca | 2084.08 |
| pesquero | 1439.24 |
| pez | 929.99 |
| población | 877.85 |
| conservación | 786.68 |
| ordenación | 776.59 |
| pabellón[6] | 590.73 |
| transzonal | 455.16 |
| pesquería | 407.25 |
| subregional | 406.26 |
| buque | 405.58 |
| captura | 401.17 |
| migratorio | 397.76 |
| recurso_pesquero | 369.23 |
| mar | 320.63 |
| mediterráneo | 319.32 |
| enarbolar[6] | 261.46 |
| tac | 232.24 |
| cgpm | 201.48 |
| flota | 192.98 |
| pescador | 184.09 |
| pescar | 181.22 |
| gestión | 173.65 |
| regional | 166.64 |
| convención | 162.90 |
| marino | 151.92 |

At the end of the training phase, long lists of associates and their keyness should exist for each descriptor in each of the languages for which we want to be able to assign EUROVOC descriptors to texts. The training phase happens once off-line. The system that actually assigns EUROVOC descriptors to new texts only accesses the existing associate lists.

### 3.2.2. Choice of reference corpus

At this stage, it seems useful to explain the impact of the choice of the reference corpus used in the application. For general-purpose applications, it is best to use a balanced corpus such as the *British National Corpus* or a large collection of newspaper articles because it will contain word frequency information for words from many different subject domains. Should the keyword identification tool then be asked to assign descriptors to a non-specialised text such as a normal newspaper article, the performance is best.

---

[5] "los buques que enarbolen pabellón de un estado miembro"

However, specific applications may require that the reference corpus consist of texts from a specific subject domain. This is the case when identifying descriptor associates using our training material because the training material pertains to some very specific sublanguages. Both the texts from the *European Parliament* and from the *EC's Publications Office* are of a legalistic or administrative nature. If the descriptor associates were calculated using a word frequency list of a general purpose reference corpus, the lists of associates for most descriptors would contain legal and administrative terms because, compared to general language, these terms *are* indeed salient. This means that in the mega-documents produced on the basis of the texts indexed with a certain EUROVOC descriptor, these legalistic terms *are* more frequent than in general language.

To avoid the bias towards this legalistic and administrative language, we use the whole collection of our training material as a reference corpus. By doing this, we get reference lists with higher frequency counts for legal and administrative words and we therefore lower the likelihood that these administrative terms become part of many of the lists of associates. By doing this, we still allow these terms to be part of the associate lists of descriptors which *are* from the legal or administrative domains. Note that excluding these sublanguage-specific terms altogether by adding them to the stop word list would have the negative consequence that they could not even be part of the associate lists of descriptors when they should.

### 3.2.3. Assignment Phase

During the assignment phase of the process, a lemma frequency list of a new document to which EUROVOC descriptors should be assigned is matched against a database with all descriptor associates of the same language. In the current implementation, the algorithm is the following: Going through the whole lemma frequency list of the document, each time a lemma is an associate of one or more descriptors, the frequency of the lemma is multiplied by the keyness value of the associate. The result is added to the overall score of the descriptor. For instance, if the word 'pesca' is found three times in a document, 3*2084 = 6252 (see Table 2) is added to the score of the descriptor FISHERY MANAGEMENT. One and the same lemma often is an associate of several descriptors, but their keyness will differ. The more associates are found in the lemma frequency list of the new document, and the higher the keyness values of the associates are, the higher will be the score for the descriptor. When the whole list of lemmas of the new document has been processed, the descriptors can be ranked by their score and they can be displayed to the user. As each EUROVOC descriptor has exactly one translation in each language, displaying the descriptors in different languages is only a matter of a database lookup. **Table 3** shows the English assignment results for a Spanish administrative text, i.e. a question to the European Parliament regarding the raising number of events of plutonium smuggling in the EU.

## 4. Evaluation

### 4.1. A note on the comparison of automatic and manual assignment results

Choosing appropriate keywords for a given text is a highly conceptual task for which it is difficult to apply clear-cut rules. For this reason, manual keyword assignment results differ from one human indexer to the other, and supervisors of human indexers tell us that their daily mood also influences the results. The assignment results furthermore differ for one and the same indexer over time as the view of the relevance of certain subject areas changes. For these reasons, manual indexing should not be taken as an absolute benchmark for automatic indexing, but lacking other alternatives, this is what we have to do. We should keep in mind, though, that there are no absolute rules which say that a certain descriptor is right or wrong for a given text.

### 4.2. Discussion of the assignment results

Table 3 shows both the manual and the automatic EUROVOC indexing results for the Spanish text. The manually assigned ones are underlined. The aim of the EP's professional (human) indexers is to identify very few highly relevant descriptors, while the number of descriptors identified by our system is parametrisable. The table shows that ten out of the eleven manually assigned descriptors were found among the first 23 suggestions of our system. The only descriptor which was not found directly (PLUTONIUM) is a *narrower term* (a hyponym) of RADIOACTIVE MATERIALS, which was found (rank 8, score 29). It is noteworthy that, if we had searched for the exact wording of the descriptor terms in the text, we would have only found

**Table 3** 23 top-scoring English Eurovoc descriptors and their score assigned automatically to the Spanish policy document *Postura de la Unión Europea frente al descubrimiento del contrabando de plutonio* (*Attitude of the European Union towards the discovery of plutonium smuggling*) (383 words long)

| Rank | Score | English Descriptor |
|---|---|---|
| 1 | 97 | <u>NUCLEAR SAFETY</u> |
| 2 | 62 | <u>NUCLEAR NON-PROLIFERATION</u> |
| 3 | 43 | <u>NUCLEAR FUEL</u> |
| 4 | 42 | *NUCLEAR POWER STATION* |
| 5 | 38 | *NUCLEAR TEST* |
| 6 | 34 | <u>IAEA</u> |
| 7 | 32 | *RADIOACTIVE WASTE* |
| 8 | 29 | <u>RADIOACTIVE MATERIALS</u> |
| 9 | 28 | *NUCLEAR ENERGY* |
| 10 | 25 | <u>ILLICIT TRADE</u> |
| 11 | 21 | *DECOMMISSIONING OF POWER STATIONS* |
| 12 | 18 | *EAEC* |
| 13 | 18 | <u>ORGANIZED CRIME</u> |
| 14 | 17 | *CIS* |
| 15 | 16 | <u>EUROPOL</u> |
| 16 | 14 | NUCLEAR ACCIDENT |
| 17 | 14 | ~~BUDGETARY DISCHARGE~~ |
| 18 | 13 | UKRAINE |
| 19 | 13 | <u>CIS COUNTRIES</u> |
| 20 | 12 | TRANSPORT OF DANGEROUS GOODS |
| 21 | 12 | ~~RESEARCH AND DEVELOPMENT~~ |
| 22 | 12 | ~~EC GENERAL BUDGET~~ |
| 23 | 11 | <u>POLICE COOPERATION</u> |
| … | | |

<u>Underlined</u>: manually assigned descriptors
*Italics*: further 'reasonable' descriptors
Normal: wrong, but semantically related
~~Strikethrough~~: wrong descriptors, semantically not related

EUROPOL and PLUTONIUM with 4 occurrences each and IAEA and CIS with one occurrence each.

Table 3 also shows that the automatically identified descriptors are not the ten highest-ranking ones. Instead, they are in positions 1, 2, 3, 6, 8, 10, 13, 15, 19 and 23. However, according to our own judgement, many more descriptors suggested by the system are also relevant for the given document (displayed in italics), and others are semantically related even if they are not entirely correct descriptors for this text (displayed in normal, non-italic font). Professional indexers would probably not consider this last group appropriate for the given text. Three of the suggested descriptors are clearly not related to the contents of the text (marked with strikethrough) so that they may actually give a wrong idea of the document's contents.

### 4.3. Recall and precision

We only completed our tool which carries out an automatic evaluation of assignment results a few days before this paper had to be submitted so that we had no time to experiment with different assignment procedures to optimise the results. We believe therefore that the assignment results will soon be much better than those shown in **Table 4** because we intend to experiment with the various parameter settings and the assignment algorithm, and we will improve the stop word and multi-word term lists used (see section 2.2).

**Table 4** Evaluation of first Eurovoc descriptor assignment results on the whole of the Spanish training corpus.

| Rank | Correct descriptor found | | Correct descriptor + RT + BT + NT found | |
|---|---|---|---|---|
| | Recall | Precision | Recall | Precision |
| 1 | 9% | 62% | 13% | 88% |
| 2 | 15% | 54% | 19% | 65% |
| 3 | 21% | 47% | 25% | 57% |
| 5 | 28% | 39% | 33% | 46% |
| 7 | 34% | 33% | 40% | 39% |
| 10 | 40% | 27% | 47% | 32% |
| 15 | 47% | 21% | 54% | 24% |
| 20 | 52% | 17% | 60% | 20% |
| 25 | 56% | 15% | 64% | 17% |
| 30 | 58% | 13% | 67% | 15% |
| 50 | 66% | 9% | 75% | 10% |
| 100 | 73% | 5% | 83% | 5% |
| Varying | 37% | 36% | 43% | 43% |

In the only experiment carried out so far we assigned the descriptors to the texts of the training corpus. It is to be expected that these results are better than results on new texts, which were not part of the training collection.

As our system assigns a ranked list of keywords to a text, recall and precision can be calculated for a number of automatically suggested descriptors we can decide on. The number of descriptors used for the evaluation is shown in the first column of the table ('rank'). The second and third columns give recall and precision measures concerning the manually assigned descriptors, which were found in the automatic procedure.

As EUROVOC is structured hierarchically (*Broader Terms* and *Narrower Terms*) and has explicit links between different parts of the thesaurus (*Related Terms*), it is reasonable to consider these structural relationships in the performance evaluation. In the results shown in columns four and five of Table 4, we considered the keyword assignment to be successful when our system found either one of the manually assigned terms, or a RT of a manually assigned descriptor, or a BT or a NT at a level *immediately* above or below the manual descriptor. With other words: when a manually assigned descriptor was not found, but a descriptor was found which is a RT to this manual descriptor, we considered the descriptor to be found. This means that, in 88% of all documents, our system either found a manually assigned descriptor (62% of all cases) or it found a RT, BT or NT of a manually assigned descriptor. If one of the 2870 descriptors had been assigned arbitrarily, precision would have been 0.035%.

The last row ('rank = varying') shows the results of an experiment in which, for each text, we considered exactly the number of manually assigned descriptors when calculating recall and precision. This number usually varies between one and twenty-five, with an average of 6.91.

Depending on the intended usage of the system, the current results are already acceptable and useful because the suggested lists do give users an idea of the approximate contents of the document, and they do certainly allow automatic document comparison (see 1.2). For document comparison, optimal performance is less important than consistency of the results. This means that if two documents about plutonium smuggling, for instance, were both indexed incorrectly with the term NUCLEAR ACCIDENT, our similarity calculation tool will recognise the two documents as similar, even if its decision is based on incorrectly assigned EUROVOC descriptors.

## 5. *Challenges and difficulties*

The automatic training of our system was complicated by some specific features of the corpus and by the way EUROVOC is used.

We mentioned earlier that the training corpus pertains to a very specific legalistic and administrative **sublanguage**. The problem is aggravated by the fact that the training material consists of different kinds of sublanguage texts (legal texts, communications, questions to the European Parliament, etc.) which each have their own typical vocabulary. When some descriptors were used more to index texts from one subcorpus and others to index the texts from another, the lists of associates are biased by the respective sublanguages. We have to investigate whether we can solve this problem by training the system on different subsets of texts, always using the appropriate subcorpus to produce the reference lemma frequency lists (see 3.2.2).

Another problem is caused by the fact that the **descriptor usage is very uneven**: some of them are used thousands of times, whereas others have been used rarely or not even once. We cannot produce associate lists of reasonable quality for those descriptors for which not enough training material is available, which means that we shall not be able to assign this descriptor to a text. However, this is a minor problem because the reason for the rare usage of this descriptor is the fact that it is not particularly useful. For instance, some descriptors are needed to provide a link in the hierarchy of descriptors without themselves being particularly useful as keywords for a text.

The varying size of the training material also has the effect that **some associate lists are much longer than others**, which means that the descriptors with long associate lists are more likely to be assigned than the descriptors with less associates. As some descriptors were used much more than others, this phenomenon may exactly yield the wanted results, i.e. the frequently used descriptors will be assigned more frequently, but this cannot be taken for granted until we have tested the results thoroughly.

Another difficulty is caused by the fact that some descriptors co-occur very frequently, the extreme being that they were always used as a pair. For these descriptors it will, of course, be rather difficult to identify individual associate lists.

Finally, we would like to point out a restriction, which applies to all instances of **controlled vocabulary indexing**, be they manual or automatic. While controlled vocabulary indexing has some clear advantages (see 1.2), its disadvantage is that (human and mechanical) indexers are limited to using these terms. To give an example, EUROVOC does not cover the field of *computational linguistics* at all and even the field of computer science is rather badly represented. This means that trying to index a text like the one you are currently reading with EUROVOC is a frustrating exercise for the indexer.

## 6. Some implementation details

The current prototype system is implemented in PERL, with the exception of the monolingual keyword assignment tool and the lemmatiser. The individual components are linked in batch mode using a PERL wrapper. The lists of descriptor associates are stored in an MS-Access database, which is accessed from within the PERL program. We have not yet invested any time in providing an interface for the application and an interface is not strictly necessary because the tool will be integrated with other tools and the user should only see the resulting Eurovoc descriptor lists.